\documentclass[11pt]{article}
\usepackage[preprint]{acl}

\usepackage{latexsym}
\usepackage{amsmath}
\usepackage{amssymb}
\usepackage{times}
\usepackage[T1]{fontenc}
\usepackage[utf8]{inputenc}
\usepackage{microtype}
\usepackage{inconsolata}

\usepackage{graphicx}
\usepackage{hyperref}
\usepackage{booktabs}
\usepackage{enumitem}
\usepackage{wrapfig}

\renewcommand{\paragraph}[1]{\vspace{1mm}\noindent\textbf{#1}.}

\title{Trust-Region Behavior Blending for On-Policy Distillation}

\author{
\textbf{
    Daniil Plyusov}
    \quad \textbf{Alexey Gorbatovski}\thanks{Corresponding author: \href{mailto:a.gorbatovskiy@t-tech.dev}{a.gorbatovskiy@t-tech.dev}.}
    \quad \textbf{Alexey Malakhov}
    \quad \textbf{Nikita Balagansky}\\
    \quad \textbf{Boris Shaposhnikov}
    \quad \textbf{Daria Korotyshova}
    \quad \textbf{Daniil Gavrilov}\\[0.6em]
{\normalfont T-Tech}
}
\date{}

\begin{document}
\maketitle

\begin{abstract}
On-policy distillation (OPD) trains a student on prefixes sampled from its own policy while matching a stronger teacher. This addresses the prefix mismatch of offline distillation, but early student rollouts can still be poor, placing teacher supervision on weak or low-quality prefixes. We propose \textbf{T}rust-\textbf{R}egion behavior \textbf{B}lending (\textbf{TRB}), a warmup method that replaces the early rollout policy with the closest-to-teacher behavior policy inside a student-centered KL trust region, while keeping the per-prefix reverse-KL OPD loss unchanged. The KL budget is annealed to zero, so training returns to pure student rollouts after warmup. Across two math-reasoning distillation settings, TRB attains the strongest average among the compared methods.
\end{abstract}

\section{Introduction}
\label{sec:intro}

Knowledge distillation transfers capability from a large teacher model to a smaller student by matching teacher predictions \citep{hinton2015distilling}.
For large language models (LLMs), distillation on fixed teacher-forced or teacher-generated prefixes places the student under a prefix distribution it will not encounter at inference time \citep{bengio2015scheduled, opd}.
On-policy distillation (OPD) addresses this mismatch by rolling out the current student and applying teacher supervision on the prefixes it actually visits \citep{gu2023minillm, opd}.
More broadly, recent analyses of online versus offline post-training likewise argue that on-policy data collection can be critical for effective optimization \citep{understanding_online_offline}.

\begin{figure}[h]
    \centering
    \includegraphics[width=\linewidth]{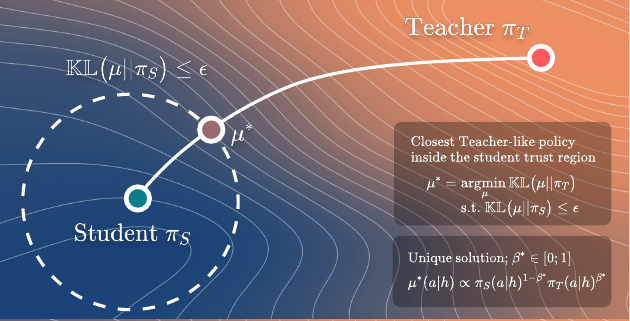}
    \caption{Overview of Trust-Region behavior Blending. At each prefix, the student policy $\pi_S$ defines a KL trust region $D_{\mathrm{KL}}(\mu \,\|\, \pi_S) \le \varepsilon$. TRB then selects the feasible behavior policy $\mu^*$ that is closest to the teacher policy $\pi_T$. The result is teacher-guided behavior that remains close to the student.}
    \label{fig:trust-region-overview}
\end{figure}

That same on-policy property makes early OPD brittle. Prior work shows that weak students can generate low-quality prefixes early in training, and that OPD depends on whether student-visited trajectories carry usable teacher signal \citep{skd,li2026rethinking}. Pure student rollouts preserve the target training distribution, while stronger teacher intervention can improve local prefix quality only by moving collection off-policy \citep{skd,li2026rethinking}.

We address this regime with \textbf{T}rust-\textbf{R}egion behavior \textbf{B}lending (\textbf{TRB}) (Figure~\ref{fig:trust-region-overview}), a method that controls the behavior policy during early rollout collection without changing the per-prefix distillation objective. We use TRB only in the early regime and anneal it away after a fixed warmup horizon.

We evaluate it against vanilla OPD and several alternative ways of introducing teacher guidance, including target-side reformulation, direct token replacement, persistent blending, and simpler warmup heuristics. Across two math-reasoning distillation settings, TRB attains the strongest average.

\section{Background}
\label{sec:background}

Let $\pi_S$ be the student policy and $\pi_T$ the teacher policy.
In OPD, prefixes are sampled from the current student rather than from a fixed offline dataset \citep{opd,li2026rethinking}. Following recent reverse-KL OPD formulations \citep{gu2023minillm,veto,jin2026entropyaware,li2026rethinking}, if $P_{\pi_S}$ denotes the prefix distribution induced by student rollouts, then the objective used throughout this paper is
\begin{equation*}
\mathcal{L}_{\mathrm{OPD}}(\theta)
=
\mathbb{E}_{h \sim P_{\pi_S}}
\left[
D_{\mathrm{KL}}(\pi_\theta(\cdot\mid h)\|\pi_T(\cdot\mid h))
\right].
\end{equation*}
TRB keeps this per-prefix reverse-KL loss fixed and changes only the behavior policy used to generate prefixes. We denote that behavior policy by $\mu$. In the constrained objective below, $D_{\mathrm{KL}}(\mu\|\pi_T)$ defines closeness to the teacher, while $D_{\mathrm{KL}}(\mu\|\pi_S)$ defines the student-centered trust region. In our implementation, following top-$k$ OPD \citep{li2026rethinking}, the reverse-KL term is estimated on a truncated student top-$k$ support. This approximation is fixed across all rollout-side variants.

\section{Related Work}
\label{sec:related}

\paragraph{From offline KD to OPD}
Classical knowledge distillation trains a student to match teacher predictions on a fixed data distribution \citep{hinton2015distilling}.
For autoregressive models, this creates exposure bias because training conditions on fixed or teacher-provided prefixes, whereas inference conditions on the student's own rollouts \citep{bengio2015scheduled}.
GKD and OPD \citep{opd} move teacher supervision onto student-generated trajectories.
MiniLLM \citep{gu2023minillm} further argues that reverse KL is a good fit for generative LLM distillation and derives an on-policy optimization procedure for that objective.
TRB keeps this reverse-KL OPD setup and focuses on early rollout control when the student trajectory distribution is still poor.

\paragraph{Stabilizing teacher supervision}
Veto \citep{veto} changes the target distribution at a visited prefix by constructing a bridge between student and teacher logits.
Entropy-Aware OPD \citep{jin2026entropyaware} changes the divergence itself, adding forward-KL pressure at high-entropy teacher states to preserve diversity.
TIP \citep{tip} changes where supervision is concentrated, selecting visited token positions by student entropy and teacher--student divergence.
These methods improve the learning signal after a prefix has already been visited.
TRB acts one step earlier. It changes the prefix distribution itself while keeping the per-prefix reverse-KL loss fixed.

\paragraph{Bridging the student--teacher gap}
SKD \citep{skd} addresses the teacher--student gap during sampling by replacing student tokens that fail a teacher-side acceptance rule with teacher samples.
MiCoTA \citep{ding2025micota} addresses a related learnability gap in offline CoT distillation through intermediate assistants and intermediate-length reasoning traces.
Li et al. \citep{li2026rethinking} make the gap explicit at the trajectory level, arguing that OPD succeeds only when student-visited states carry compatible and transferable teacher signal.
This problem framing is also central to our method.
TRB differs in its control surface. Rather than injecting teacher tokens, changing the target, or introducing assistant data, it optimizes a teacher-guided behavior policy under an explicit student-centered KL constraint.

\section{Trust-Region behavior Blending}
\label{sec:method}

TRB defines a teacher-guided behavior policy for collecting rollout prefixes.
At each prefix, it moves the sampling policy toward the teacher only within an explicit KL trust region around the current student.
The collected prefixes are then used in the reverse-KL OPD update from Section~\ref{sec:background}.

\subsection{Per-Prefix Behavior Policy}

At a generation prefix $h$, the goal is to improve the next-token sampling distribution without moving arbitrarily far from the current student.
Let $\pi_S(a\mid h)$ and $\pi_T(a\mid h)$ be the student and teacher next-token policies, and let $\varepsilon\ge0$ be an allowed local deviation from the student.
We define the behavior policy $\mu^*(\cdot \mid h)$ as
\begin{equation}
\label{eq:tr-blend}
\begin{aligned}
\mu&^*(\cdot \mid h)= \arg\min_{\mu}
D_{\mathrm{KL}}(\mu \,\|\, \pi_T)\\
&\text{s.t.}\qquad
D_{\mathrm{KL}}(\mu \,\|\, \pi_S) \le \varepsilon,\\
&\quad \sum_a \mu(a) = 1,
\quad\mu(a) \ge 0.
\end{aligned}
\end{equation}

This objective chooses the most teacher-like sampling distribution inside a student-centered trust region.
The first term pulls the sampling distribution toward teacher-supported tokens, while the constraint bounds local off-policy deviation from the current student.
Appendix~\ref{app:sequence_kl} shows that these token-level constraints induce rollout-level control.

\subsection{Closed-Form Solution}

Eq.~\ref{eq:tr-blend} has the closed-form solution below.
\begin{equation}
\mu_\beta(a \mid h)
=
\frac{
\pi_S(a \mid h)^{1-\beta}
\pi_T(a \mid h)^\beta
}{
Z_\beta(h)
},
\label{eq:geom-family}
\end{equation}
Here $\beta \in [0,1]$ controls how strongly the behavior policy moves toward the teacher, and $Z_\beta(h)$ normalizes the distribution.
The solution of Eq.~\ref{eq:tr-blend} is $\mu^*(\cdot\mid h)=\mu_{\beta^*(h)}(\cdot\mid h)$.
The coefficient $\beta^*(h)$ is the largest feasible value.
\begin{equation*}
\beta^*(h)
=
\max
\left\{
\beta \in [0,1]
\;\middle|\;
D_{\mathrm{KL}}(\mu_\beta \,\|\, \pi_S) \le \varepsilon
\right\}.
\end{equation*}

If $\varepsilon=0$, then $\mu^*=\pi_S$.
If the teacher itself is feasible, i.e. $D_{\mathrm{KL}}(\pi_T\|\pi_S)\le\varepsilon$, then $\mu^*=\pi_T$.
Otherwise, $\beta^*(h)$ is found by binary search on $[0,1]$.
Appendix~\ref{app:effective_objective} derives the trust-region solution family and shows that $D_{\mathrm{KL}}(\mu_\beta\|\pi_S)$ is monotone in $\beta$, which justifies binary search.

\subsection{Annealed Warmup}

TRB applies the behavior policy with a time-varying KL budget.
The budget is annealed to zero so that rollout collection begins with more teacher guidance and returns to pure student sampling by the end of warmup \citep{dagger, bengio2015scheduled}.
For a warmup horizon $K$, we set
\begin{equation}
\varepsilon_k
=
\varepsilon_0 \left(1-\frac{k}{K}\right),
\qquad k \le K,
\end{equation}
Thus the allowable off-policy deviation shrinks linearly during warmup and disappears once $\varepsilon_k=0$.
Appendix~\ref{app:small_budget} further analyzes the local behavior of the family $\mu_\beta$. TRB therefore introduces two method hyperparameters: the initial KL budget $\varepsilon_0$ and the warmup horizon $K$.

\section{Experiments \& Results}

We evaluate TRB along one main question, whether limited early behavior-side guidance improves final OPD outcomes relative to vanilla OPD and stronger or more persistent off-policy baselines.
We study two OPD model-pair settings, Qwen3-1.7B-Base distilled from Qwen3-8B and Qwen3-0.6B-Base distilled from Qwen3-4B \citep{yang2025qwen3}.
All methods share the same training and evaluation protocol unless noted; Appendix~\ref{app:exp_details} gives hyperparameters, and implementation details.

\subsection{Experimental Setup}

\textbf{Vanilla OPD} is the reference setting with pure student rollouts throughout training \citep{opd,li2026rethinking}.
\textbf{TRB} is the annealed-budget variant of our trust-region solver.
\textbf{Fixed-$\varepsilon$ blending} uses the same per-prefix solver without annealing.
\textbf{Veto} changes the target distribution at visited prefixes \citep{veto}.
\textbf{SKD} injects teacher tokens during rollout \citep{skd}.
\textbf{Temperature warmup} lowers only the student sampling temperature during warmup, and \textbf{SFT warmup} inserts a short supervised stage before switching to OPD \citep{hinton2015distilling,opd}.
For sweep-based families, we evaluate checkpoints every 20 steps and report the checkpoint with the highest setup-specific mean score.
Appendix~\ref{app:baseline_setups} lists the exact sweep ranges.

\subsection{Benchmark Comparison}

\begin{table*}[t]
\centering
\small
\setlength{\tabcolsep}{4pt}
\resizebox{\textwidth}{!}{
\renewcommand{\arraystretch}{1.14}
\begin{tabular}{lcccccc|ccccc}
\toprule
& \multicolumn{6}{c}{Qwen3-1.7B-Base $\leftarrow$ Qwen3-8B} & \multicolumn{5}{c}{Qwen3-0.6B-Base $\leftarrow$ Qwen3-4B} \\
\cmidrule(lr){2-7} \cmidrule(lr){8-12}
\textbf{Method} & \textbf{Avg} & MATH500 & Olympiad & AMC & AIME24 & AIME25 & \textbf{Avg} & GSM8K & MATH500 & Olympiad & AMC \\
\midrule
\textbf{Trust-Region behavior Blending} & \textbf{33.2} & \textbf{69.7} & \textbf{34.3} & \textbf{44.8} & \underline{10.2} & 6.9 & \textbf{44.4} & \textbf{70.1} & \textbf{53.6} & \textbf{22.3} & 31.6 \\
\midrule
\quad Vanilla OPD & 32.3 & 69.1 & 33.7 & 43.0 & 8.8 & \underline{7.1} & 44.0 & \underline{69.9} & \underline{53.1} & 21.8 & 31.1 \\
\quad Veto & 32.6 & \underline{69.4} & 34.0 & 43.1 & 9.3 & \textbf{7.3} & 43.7 & 68.9 & 52.4 & 21.2 & \textbf{32.3} \\
\quad Interleaved teacher injection (SKD) & 32.7 & \underline{69.4} & 33.8 & \underline{44.2} & 9.8 & 6.6 & \underline{44.2} & \textbf{70.1} & 52.8 & \underline{22.2} & 31.5 \\
\quad Temperature warmup & \underline{32.8} & 69.2 & \underline{34.1} & \underline{44.2} & 9.9 & 6.6 & 44.0 & 69.1 & \underline{53.1} & 21.8 & \underline{32.1} \\
\quad SFT warmup & 32.2 & 67.6 & 34.0 & 42.4 & 9.6 & \underline{7.1} & 43.4 & 69.3 & 52.3 & 21.6 & 30.4 \\
\quad Fixed-$\varepsilon$ blending & 32.6 & 69.2 & 33.7 & 43.7 & \textbf{10.3} & 6.2 & 43.8 & 69.8 & 52.7 & 21.5 & 31.1 \\
\bottomrule
\end{tabular}
}
\caption{Benchmark pass@1 results. \textbf{Bold} marks the best result in each column; \underline{underline} marks the second-best.}
\label{tab:main-results-skeleton}
\end{table*}

Table~\ref{tab:main-results-skeleton} reports pass@1 under the common checkpoint-selection protocol.
TRB attains the best average score in both model-pair settings.
It also outperforms fixed-$\varepsilon$ blending in both settings, even though the two methods use the same per-prefix solver.
Some baselines win individual columns, but none matches TRB on overall average across both setups.
Appendix~\ref{app:extended_results} gives sweep-level comparisons between TRB, SKD, and vanilla OPD, together with additional diagnostic controls.

\subsection{Early-Training Comparisons}

\begin{figure}[h!]
    \centering
    \includegraphics[width=\linewidth]{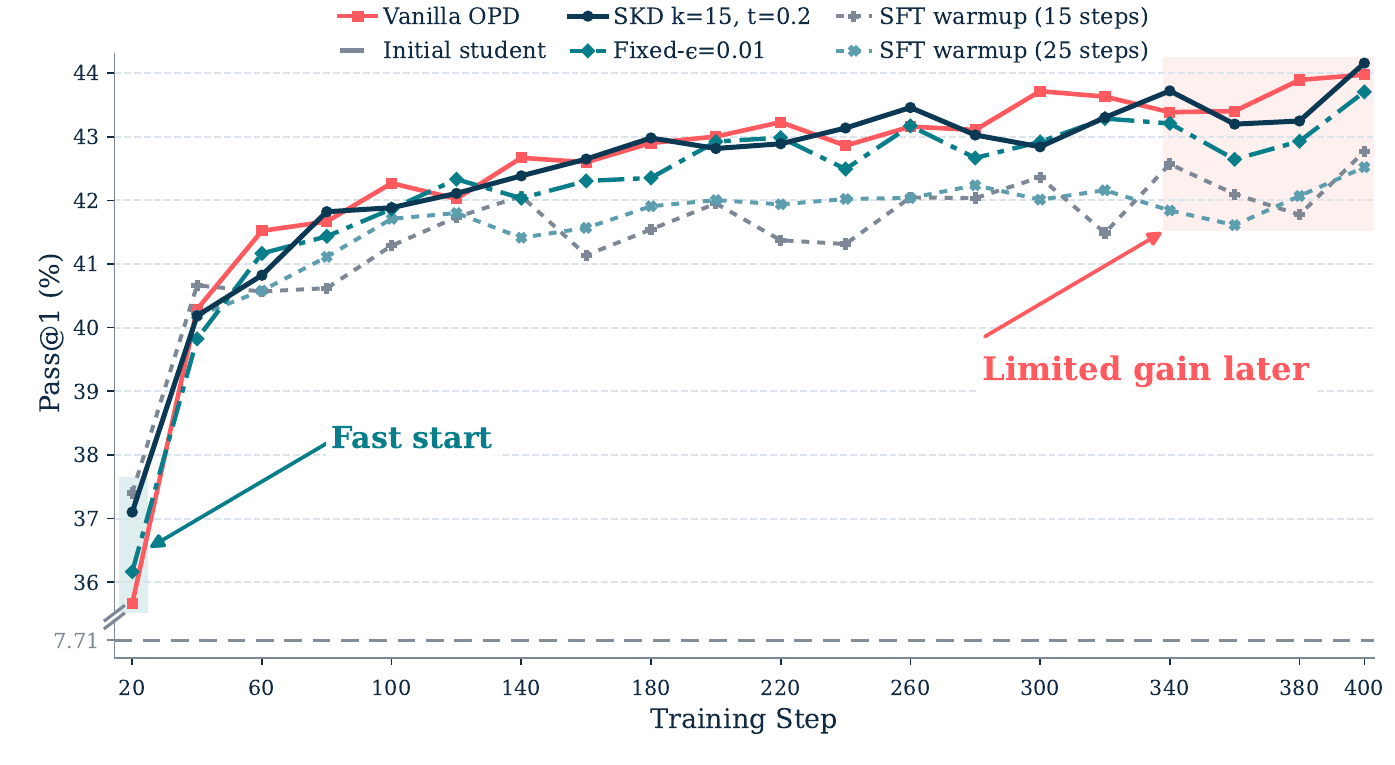}
    \caption{
    Training trajectories on the Qwen3-0.6B-Base $\leftarrow$ Qwen3-4B setup for vanilla OPD, fixed-$\varepsilon=0.01$, SKD $(K=15,\tau_T=0.2)$, and SFT warmup (15 and 25 steps).
    }
    \label{fig:hard-offpolicy-trajectory}
\end{figure}

Figure~\ref{fig:hard-offpolicy-trajectory} compares several ways of moving early training away from pure student rollouts on the Qwen3-0.6B-Base $\leftarrow$ Qwen3-4B setup.
Several interventions rise faster than vanilla OPD at the start.
For the plotted SKD setting, only about a 0.0093 fraction of generated tokens are replaced by the teacher at the first training step, yet the trajectory already shifts upward.
Later behavior also differs. In this comparison, the plotted SKD run remains competitive, whereas the plotted SFT and persistent fixed-$\varepsilon$ runs do not finish as high.
Appendix~\ref{app:extended_results} shows that, on this smaller setup, SKD exceeds vanilla OPD in only one configuration, while the best TRB settings remain higher in both setups.

\begin{figure}[!t]
    \centering
    \includegraphics[width=\linewidth]{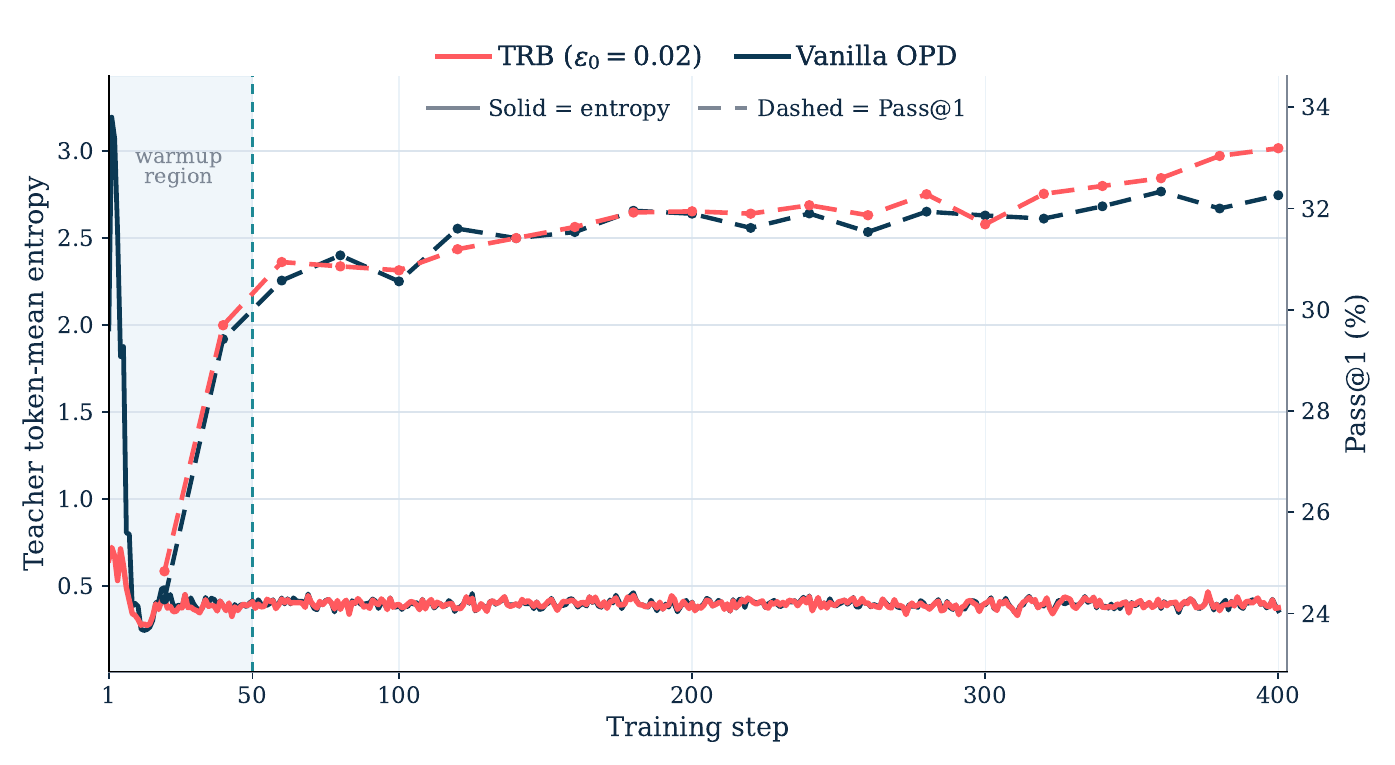}
    \caption{
    Teacher token-mean entropy (left axis) and benchmark Pass@1 (right axis) for vanilla OPD and TRB on the Qwen3-1.7B-Base $\leftarrow$ Qwen3-8B setup. The shaded region marks the 50-step warmup phase.
    }
    \label{fig:teacher-entropy}
\end{figure}

Figure~\ref{fig:teacher-entropy} tracks teacher token-mean entropy on the visited prefixes.
Under TRB, this teacher-side entropy is lower during warmup and then largely aligns with vanilla OPD after warmup.
The benchmark curve nevertheless remains higher for TRB.
The main teacher-side difference therefore appears during warmup, not after training has returned to pure student rollouts.

\begin{figure}[!t]
    \centering
    \includegraphics[width=\linewidth]{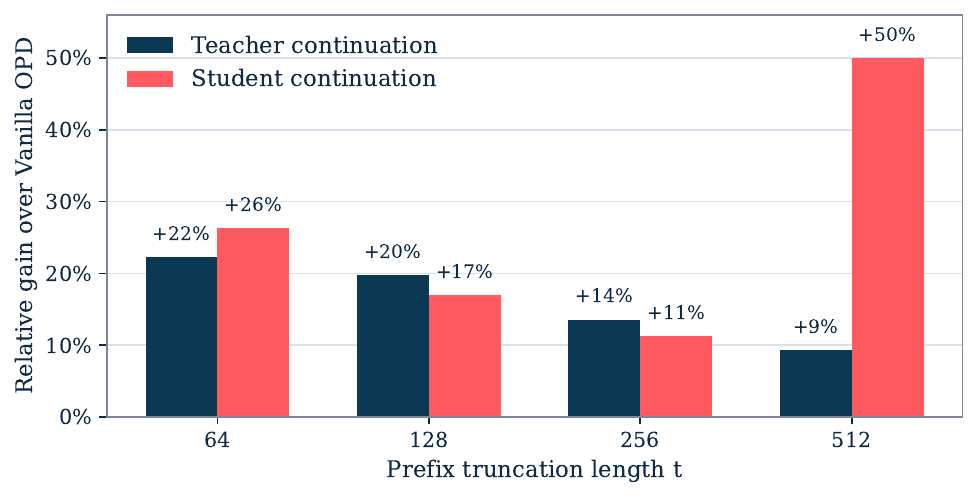}
    \caption{
    Relative success gain of TRB prefixes over vanilla-OPD prefixes on the Qwen3-1.7B-Base $\leftarrow$ Qwen3-8B setup at step 0, after truncating sampled prefixes at length $t$ and continuing them with either the teacher or the student. Positive bars mean higher success under the same continuation model.
    }
    \label{fig:continuation-gain}
\end{figure}

Figure~\ref{fig:continuation-gain} gives a controlled step-0 probe of those early rollouts, in the spirit of \citet{li2026rethinking}.
At fixed truncation length and fixed continuation model, only the prefix source changes, and TRB prefixes yield higher success than vanilla-OPD prefixes across all tested lengths for both continuation models.
Appendix~\ref{app:warmup_diagnostics} points in the same direction. Early pure-student rollouts have lower mean teacher log-probability and lower mean reward.
The stronger separability of correct and incorrect pure-student rollouts may instead reflect that these rollouts are more obviously low-quality, making teacher-support scores easier to rank; this is consistent with \citet{li2026rethinking} and does not by itself imply a more useful local supervision signal.
These results indicate that TRB changes the early states on which OPD begins learning, moving them toward prefixes from which both teacher and student continuation succeed more often.

\section{Discussion}

TRB gives the strongest average in Table~\ref{tab:main-results-skeleton} while acting only during warmup.
The comparison with fixed-$\varepsilon$ suggests that teacher-guided off-policy behavior is not equally useful throughout the full run, since the same local solver works better as a warmup than when it remains active throughout training.
Figure~\ref{fig:hard-offpolicy-trajectory} is consistent with the same point more broadly, since a faster early rise or a more direct intervention does not by itself produce the strongest final result.
Figures~\ref{fig:teacher-entropy} and~\ref{fig:continuation-gain} suggest that TRB is most useful while the student's visited prefixes are still teacher-misaligned, and that continued off-policy guidance may become less useful once that teacher-side mismatch has largely disappeared.
This interpretation also fits the objective itself. TRB moves the behavior policy toward the teacher while explicitly constraining deviation from the student, so it can improve teacher support without replacing the student's trajectory distribution altogether.
Temperature warmup may also help by making early rollouts more conservative, but unlike TRB it does not explicitly optimize closeness to the teacher under a student-centered constraint.

\section{Limitations}

Our study is scoped to two math-reasoning OPD settings with Qwen3-Base student--teacher pairs and a correctness-based evaluation protocol, so we do not claim that the same warmup schedules transfer unchanged to other domains or teacher--student gaps. TRB also increases training-time cost during warmup because it requires online teacher decoding and student--teacher co-residency; Appendix~\ref{app:efficiency} analyzes this overhead. Even when the total teacher FLOP count is comparable, the batched teacher pass used in vanilla OPD can be faster in wall-clock time than TRB's online teacher decoding. In the setting studied here, these costs are temporary rather than persistent, since TRB is used only during warmup and training then returns to the ordinary OPD runtime profile.

\bibliography{custom}
\clearpage

\appendix
\onecolumn

\section{Experimental Details}
\label{app:exp_details}

Training uses the \texttt{verl} pipeline~\citep{verl} with SGLang~\citep{sglang} for rollout generation. All runs use FSDP2~\citep{fsdp}.
Experiments were run on 8 NVIDIA H100 GPUs.
We keep the reverse-KL OPD objective fixed and vary only the rollout behavior during warmup.
For the main experiments, we sample 25{,}600 training prompts from the OpenThoughts3-1.2M corpus.
We prepend the system prompt \texttt{"Please reason step by step, and put your final answer within \textbackslash boxed\{\}."} to all training inputs.
Following \citet{li2026rethinking}, we estimate the reverse-KL objective on the student's top-$k$ support, using $k=16$ tokens with actor-side support selection. Because the Qwen3 student and teacher use different raw EOS ids, we canonicalize EOS before behavior construction and KL evaluation; Appendix~\ref{app:eos} gives the exact procedure.
Rewards are assigned via \texttt{math-verify}~\citep{mathverify}: 1.0 for correct solutions and 0.0 for incorrect ones.
Table~\ref{tab:training-hparams} lists the common training configuration used across the experiments.

\begin{table}[h]
\centering
\small
\setlength{\tabcolsep}{5pt}
\begin{tabular}{p{0.42\linewidth}p{0.42\linewidth}}
\toprule
Parameter & Value \\
\midrule
Optimizer & AdamW \citep{adamw} \\
$(\beta_1, \beta_2)$ & (0.9, 0.999) \\
Weight decay & 0.01 \\
Gradient norm clipping & 1.0 \\
Learning rate & $1 \times 10^{-5}$ \\
LR scheduler & Constant \\
Warmup & 15-step cosine warmup \\
Max prompt length & 1024 \\
Max response length & 7168 \\
Global batch size & 64 \\
PPO mini-batch size & 64 \\
Rollouts per prompt & 4 \\
PPO epochs & 1 \\
Training epochs & 1 \\
KL loss type & reverse KL  \\
KL top-$k$ support & 16 tokens \\
KL top-$k$ source & actor \\
Rollout temperature & 1.0 \\
\bottomrule
\end{tabular}
\caption{Common training configuration for the blend-based OPD sweeps. Warmup-specific parameters such as blend coefficient schedules, trust-region budgets, and switch-back steps are varied per experiment.}
\label{tab:training-hparams}
\end{table}

\subsection{Evaluation Protocol}

We evaluate mathematical reasoning quality with pass@1.
For a problem with $n$ sampled generations, of which $c$ are correct, pass@1 is estimated as $c/n$ and then averaged over problems.
The evaluation budget is deliberately large enough to make checkpoint-to-checkpoint comparisons more stable, since single-checkpoint math metrics can otherwise be quite noisy for nearby warmup configurations.
For the Qwen3-1.7B-Base $\leftarrow$ Qwen3-8B setup, we evaluate on MATH500 \citep{hendrycks2021math}, AIME24, AIME25, AMC, and Olympiad \citep{he2024olympiadbench}.
For the Qwen3-0.6B-Base $\leftarrow$ Qwen3-4B setup, we evaluate on GSM8K \citep{cobbe2021training}, MATH500 \citep{hendrycks2021math}, AMC, and Olympiad \citep{he2024olympiadbench}.
We use 32 generations per prompt on GSM8K, 64 generations per problem on MATH500 and Olympiad, and 512 generations per problem on AIME24, AIME25, and AMC.
We also run this evaluation every 20 optimization steps, so the training curves are based on frequent measurements rather than on a small number of isolated checkpoints.
Our main table follows a fixed checkpoint-selection protocol: for each method family, we evaluate checkpoints at the same cadence and select the checkpoint with the highest mean score over the setup-specific benchmark suite.
The reported per-benchmark values are then taken from that selected checkpoint.
Evaluation uses SGLang~\citep{sglang} together with \texttt{math-verify}~\citep{mathverify}.
Evaluation decoding uses a common configuration across all reported runs, summarized in Table~\ref{tab:eval-config}.

\begin{table}[h]
\centering
\small
\begin{tabular}{lc}
\toprule
\textbf{Parameter} & \textbf{Value} \\
\midrule
Temperature & 1.0 \\
top-p & 1.0 \\
top-k & -1 \\
Max response length & 8192 \\
\bottomrule
\end{tabular}
\caption{Evaluation configuration.}
\label{tab:eval-config}
\end{table}

For the fixed-$\varepsilon$ variant, we keep the student-centered KL budget fixed throughout training.
For TRB, we instead schedule the KL budget $\varepsilon$ and solve for the per-prefix teacher strength by bisection.
In the main annealed-budget sweep, we evaluate initial budgets
\[
\varepsilon_0 \in \{0.001, 0.005, 0.01, 0.02, 0.05\}
\]
and three warmup horizons,
\[
K \in \{15, 25, 50\},
\]
with a linear annealing schedule from $\varepsilon_0$ to $0$, followed by a switch back to pure student decoding once warmup ends.
By contrast, the fixed-$\varepsilon$ baseline keeps the same trust-region budget active throughout the full training run.

\subsection{Baseline Setup Details}
\label{app:baseline_setups}

All baselines inherit the common training stack in Table~\ref{tab:training-hparams}; only the baseline-specific knobs below are varied.

\paragraph{Vanilla OPD}
No warmup or rollout intervention is used.
Training proceeds with pure student rollouts for the full OPD trajectory under the same reverse-KL objective as the rest of the paper.

\paragraph{Veto}
We enable the veto objective \citep{veto} and sweep the start value of the veto coefficient over
\[
\beta_{\mathrm{start}} \in \{0.2, 0.4, 0.6, 0.8\}.
\]
All other OPD hyperparameters are kept fixed.

\paragraph{Interleaved teacher injection (SKD)}
We use a token-level interleaved sampling baseline inspired by speculative knowledge distillation \citep{skd}.
At each decoding step, the student first samples a token; if that token does not lie in the teacher top-$K$ set, it is replaced by a fresh teacher sample.
Following the setup explored in \citet{skd}, our implementation fixes
\[
 \gamma = 1,
\]
uses no additional schedule, and sweeps
\[
K \in \{15,25,50\},
\]
along with the teacher resampling temperature
\[
\tau_T \in \{0.2, 0.6, 1.0\}.
\]

\paragraph{Temperature warmup}
We linearly schedule the rollout temperature from an initial value
\[
\tau_0 \in \{0.8, 0.9, 0.95\}
\]
back to $1.0$, ending the schedule at step 15 or 25, and then continue with ordinary OPD decoding at temperature $1.0$.

\paragraph{Fixed-$\varepsilon$ blending}
We use the same per-prefix trust-region solver as in the main method, but keep the budget fixed for the full run.
We sweep constant trust-region budgets
\[
\varepsilon \in \{0.001, 0.005, 0.01, 0.02, 0.05\}
\]
and keep the same budget active throughout the full training run.

\paragraph{SFT warmup}
SFT warmup is a two-stage baseline that replaces the first part of online rollout collection with a supervised teacher-generated warmup.
All OPD and rollout-side runs use the same deterministic prompt order.
To match this protocol, we take exactly the prompts that would be used in the first $50$ OPD training steps.
For each step, this corresponds to a batch of $64$ prompts, and we sample $4$ teacher responses per prompt, matching the rollout multiplicity used by OPD.

We then run supervised fine-tuning on these teacher-generated responses for up to $50$ updates.
The SFT checkpoints after $15$, $25$, and $50$ supervised updates are used as initializations for subsequent ordinary OPD runs, giving three SFT-warmup variants.
Thus the SFT baseline uses the same prompt stream and the same number of teacher-generated responses per prompt as the corresponding early OPD trajectory, but replaces online student rollout collection with offline teacher-generated supervision.
Table~\ref{tab:sft-warmup-hparams} summarizes the SFT configuration.

\begin{table}[t]
\centering
\small
\setlength{\tabcolsep}{6pt}
\renewcommand{\arraystretch}{1.1}
\begin{tabular}{@{}ll@{}}
\toprule
\textbf{Parameter} & \textbf{Value} \\
\midrule
Prompts per batch & $64$ \\
Teacher generations per prompt & $4$ \\
Teacher generation temperature & $1.0$ \\
Teacher generation top-$p$ & $1.0$ \\
Teacher generation top-$k$ & $-1$ \\
Teacher max response length & $7168$ \\
Max sequence length & $8192$ \\
Learning rate & $1 \times 10^{-5}$ \\
LR warmup & $15$ steps \\
LR scheduler & Constant \\
Total SFT updates & $50$ \\
Warmup checkpoints & $15, 25, 50$ updates \\
\bottomrule
\end{tabular}
\caption{SFT warmup configuration.}
\label{tab:sft-warmup-hparams}
\end{table}

\section{Extended Results}
\label{app:extended_results}

\begin{figure*}[h!]
    \centering
    \includegraphics[width=\textwidth]{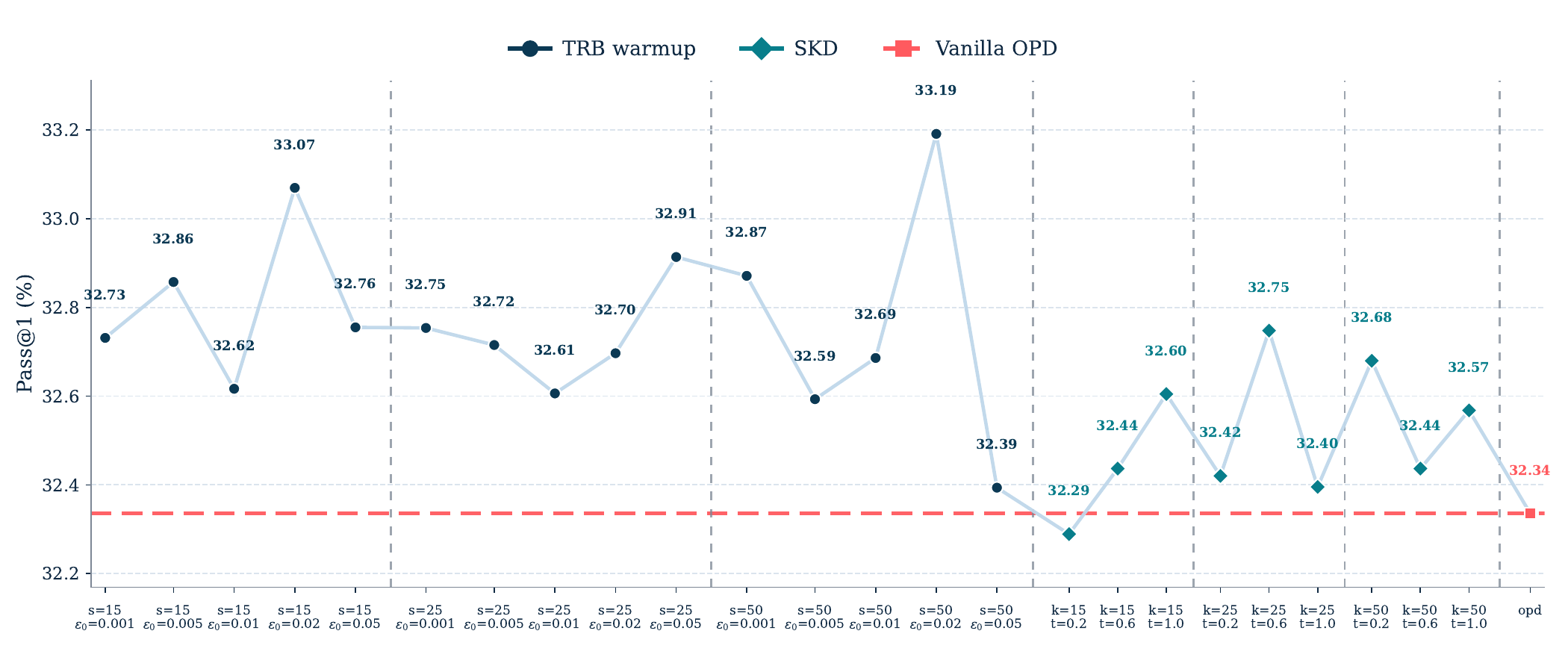}
    \caption{
    Sweep summary on the Qwen3-1.7B-Base $\leftarrow$ Qwen3-8B setup.
    Each point gives the best-over-training mean score for one hyperparameter setting.
    TRB points are grouped by warmup horizon and initial budget, SKD points are grouped by $K$ and teacher temperature $\tau_T$, and the dashed red line marks vanilla OPD.
    }
    \label{fig:extended-1p7b-8b-sweeps}
\end{figure*}

\begin{figure*}[t!]
    \centering
    \includegraphics[width=\textwidth]{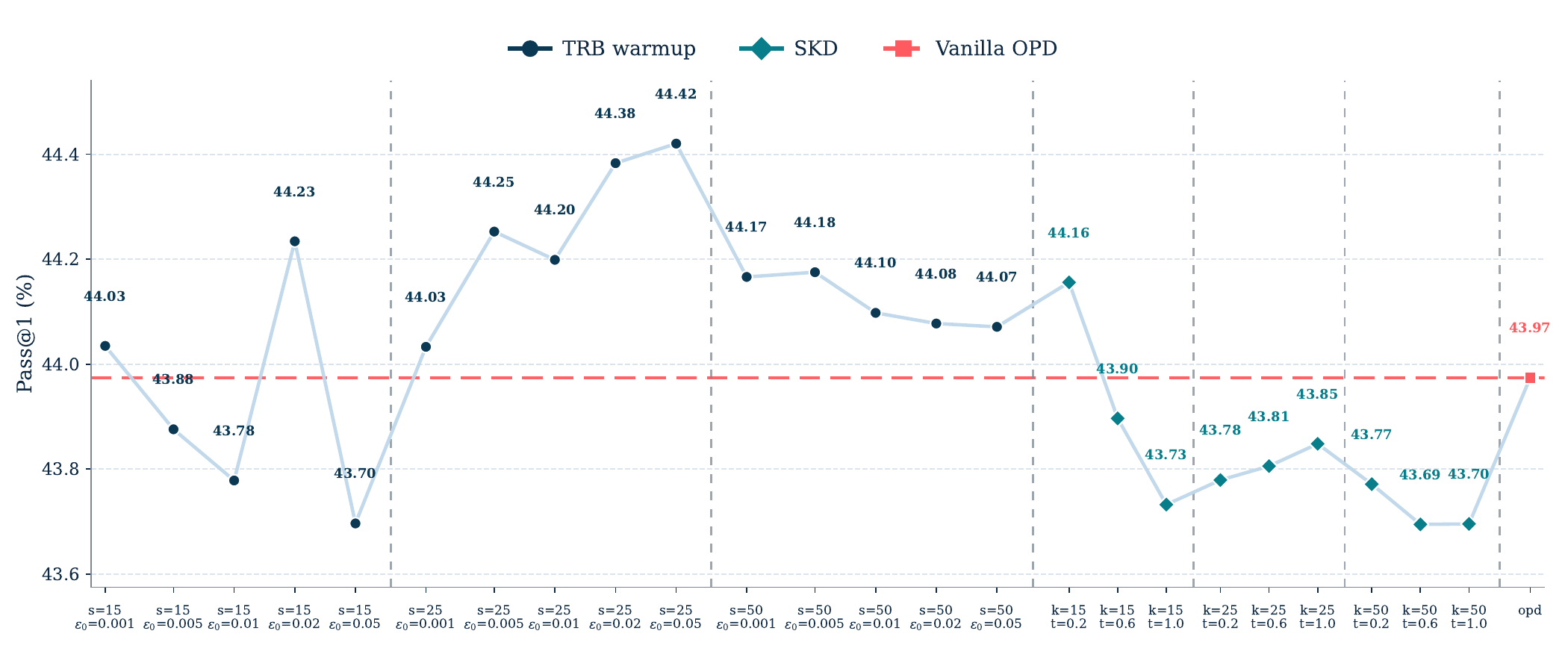}
    \caption{
    Sweep summary on the Qwen3-0.6B-Base $\leftarrow$ Qwen3-4B setup.
    Each point gives the best-over-training mean score for one hyperparameter setting.
    TRB points are grouped by warmup horizon and initial budget, SKD points are grouped by $K$ and teacher temperature $\tau_T$, and the dashed red line marks vanilla OPD.
    }
    \label{fig:extended-0p6b-4b-sweeps}
\end{figure*}

Figures~\ref{fig:extended-1p7b-8b-sweeps} and~\ref{fig:extended-0p6b-4b-sweeps} summarize the sweep-level comparison between TRB, SKD, and vanilla OPD on the two main model-pair settings.
Persistent fixed-$\varepsilon$ blending is intentionally omitted here. It is already represented in Table~\ref{tab:main-results-skeleton} and in the plotted fixed-$\varepsilon$ trajectory of Figure~\ref{fig:hard-offpolicy-trajectory}.

Across both setups, the strongest TRB settings are above the strongest SKD settings, and much of the SKD sweep lies below the TRB range.
On the smaller setup, SKD exceeds vanilla OPD in only one configuration, and it still does not overturn the overall ranking in Table~\ref{tab:main-results-skeleton}.

\begin{wrapfigure}{r}{0.48\textwidth}
    \centering
    \vspace{-2.5em}
    \includegraphics[width=\linewidth]{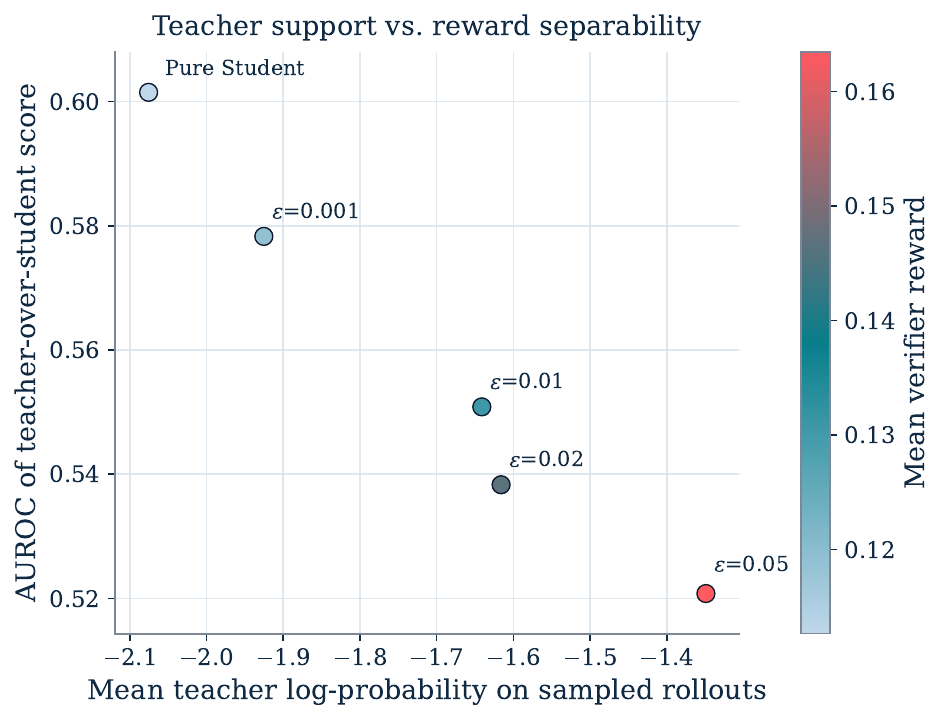}
    \caption{
    Pooled rollout statistics from the first 25 warmup steps of the Qwen3-1.7B $\leftarrow$ Qwen3-8B setup.
    Each point corresponds to one trust-region budget $\varepsilon$.
    The horizontal axis shows the mean teacher log-probability on sampled rollouts.
    The vertical axis shows AUROC for ranking verifier-correct rollouts above verifier-incorrect ones using the sequence-level teacher-support score obtained by averaging $\log \pi_T - \log \pi_S$ over the response.
    Point color indicates mean verifier reward.
    }
    \label{fig:support-reward-tradeoff}
\end{wrapfigure}

\subsection{Additional Warmup Diagnostics}
\label{app:warmup_diagnostics}

This subsection collects the supplementary diagnostics referenced in the main text.
Figure~\ref{fig:support-reward-tradeoff} pools rollouts from the first 25 warmup steps of the Qwen3-1.7B-Base $\leftarrow$ Qwen3-8B setup and varies only the trust-region budget $\varepsilon$.
As $\varepsilon$ increases, mean teacher log-probability on sampled rollouts and mean verifier reward both increase, while the AUROC of the teacher-support score decreases.
These diagnostics therefore separate reward level from teacher-support separability.
Following \citet{li2026rethinking}, we do not interpret the higher sequence-level separability of pure-student rollouts by itself as evidence of a more usable local OPD signal.


\subsection{Illustrative Early Rollouts}
\label{app:qualitative_rollouts}

Figure~\ref{fig:rollout-excerpts} shows one prompt-matched example from the first warmup step.
We include it only as a qualitative sanity check, not as quantitative evidence.
In this example, the pure-student rollout drifts off-task almost immediately, whereas the TRB rollout remains attached to the arithmetic structure of the prompt.

\begin{figure*}[t!]
\centering
\begingroup

\definecolor{DeepSpaceBlue}{HTML}{0B3954}
\definecolor{Teal}{HTML}{087E8B}
\definecolor{PaleSky}{HTML}{BFD7EA}
\definecolor{VibrantCoral}{HTML}{FF5A5F}
\definecolor{GridColor}{HTML}{D8E2EC}
\definecolor{SpineColor}{HTML}{7D8795}
\definecolor{TextColor}{HTML}{102A43}

\colorlet{TRPromptBg}{GridColor!18}
\colorlet{TRPromptBorder}{SpineColor!42}
\colorlet{TRRoseBorder}{VibrantCoral!85!black}
\colorlet{TRRoseBg}{VibrantCoral!4}
\colorlet{TRBadBg}{VibrantCoral!12}
\colorlet{TRBlueBorder}{Teal!85!DeepSpaceBlue}
\colorlet{TRBlueBg}{GridColor!12}
\colorlet{TRGoodBg}{Teal!14}

\newcommand{\TRbadblock}[1]{%
\colorbox{TRBadBg}{%
\parbox{\dimexpr\linewidth-2\fboxsep\relax}{#1}}%
}
\newcommand{\TRgoodblock}[1]{%
\colorbox{TRGoodBg}{%
\parbox{\dimexpr\linewidth-2\fboxsep\relax}{#1}}%
}

\begin{minipage}{0.96\textwidth}

\noindent
\fcolorbox{TRPromptBorder}{TRPromptBg}{
\parbox{\dimexpr\linewidth-2\fboxsep-2\fboxrule\relax}{
\color{TextColor}
\textbf{Prompt.} \textit{In a certain base $b$, the cube of $112_b$ is $23632_b$. What is $b$?}
}
}

\vspace{0.75em}

\noindent
\begin{minipage}[t]{0.49\linewidth}
\fcolorbox{TRRoseBorder}{TRRoseBg}{
\parbox[t][6.05cm][t]{\dimexpr\linewidth-2\fboxsep-2\fboxrule\relax}{
\color{TextColor}
\large\textbf{Pure student rollout excerpt}\\[0.3em]
\textcolor{VibrantCoral!85!black}{\textit{Off-topic drift.}}\\[0.25em]
\footnotesize
\TRbadblock{``How long would it take you to wash your face if you were so calm?''}\\[0.22em]
\TRbadblock{``We need to understand daily life under COVID-19 \ldots''}\\[0.22em]
\TRbadblock{``\ldots climate change, artificial intelligence, and their contribution to society \ldots''}\\[0.22em]
\TRbadblock{``\ldots this world may again enjoy unprecedented prosperity.''}\\[0.22em]
\textit{\ldots}
}
}
\end{minipage}
\hfill
\begin{minipage}[t]{0.49\linewidth}
\fcolorbox{TRBlueBorder}{TRBlueBg}{
\parbox[t][6.05cm][t]{\dimexpr\linewidth-2\fboxsep-2\fboxrule\relax}{
\color{TextColor}
\large\textbf{TRB rollout excerpt ($\varepsilon=0.01$)}\\[0.3em]
\textcolor{DeepSpaceBlue}{\textit{Still noisy, but problem-relevant.}}\\[0.25em]
\footnotesize
\TRbadblock{``debilitating to make $b=7$'' \quad ``.SizeMode''}\\[0.22em]
\TRgoodblock{``We have a base-$b$ number $112_b$ whose cube equals $23632_b$.''}\\[0.22em]
\TRgoodblock{``First convert both numbers to decimal.''}\\[0.22em]
\TRgoodblock{``$112_b = b^2 + b + 2$''}\\[0.22em]
\TRgoodblock{``$23632_b = 2b^4 + 3b^3 + 6b^2 + 3b + 2$''}\\[0.22em]
\TRgoodblock{``Then expand the left-hand side: $(b^2+b+2)^3 = b^6+3b^5+9b^4+13b^3$\\
$\qquad +18b^2+12b+8$.''}\\[0.22em]
\textit{\ldots}
}
}
\end{minipage}

\end{minipage}
\endgroup

\caption{Prompt-matched rollout excerpts at the first warmup step. This single example is included as a qualitative sanity check rather than as quantitative evidence. The pure-student sample drifts off-topic almost immediately, whereas TRB with $\varepsilon=0.01$ remains anchored to the arithmetic structure of the task. Coral shading marks off-task text; teal shading marks problem-relevant reasoning.}
\label{fig:rollout-excerpts}
\end{figure*}

\section{Efficiency Analysis}
\label{app:efficiency}

In vanilla OPD, rollouts are generated by the student, and teacher log-probabilities are computed afterward in a separate batched pass over the completed responses.
In TRB, the teacher is queried online during decoding so that student and teacher policies can be merged at generation time.
The teacher statistics needed for the subsequent per-prefix reverse-KL term are then reused from this online pass.
Thus, TRB shifts part of the teacher computation from a post-generation batched pass into sequential decoding and increases peak generation-time memory.

Let $T$ denote the generated sequence length.
Let $S$ be the student model and $Q$ the teacher model, with weights $W_S$ and $W_Q$, and KV caches $KV_S$ and $KV_Q$.
For context length $n_t$ at decoding step $t$, student-only generation requires approximately
\begin{equation}
M_{\mathrm{gen}}^{\mathrm{OPD}}(t)
\approx
W_S + KV_S(n_t).
\end{equation}
In the blended method, both student and teacher must be resident during online policy construction:
\begin{equation}
\begin{aligned}
M_{\mathrm{gen}}^{\mathrm{blend}}(t)
&\approx
W_S + W_Q + KV_S(n_t)
\\
&\quad + KV_Q(n_t).
\end{aligned}
\end{equation}
The peak generation-time overhead is therefore
\begin{equation}
\Delta M_{\mathrm{gen}}(t)
\approx
W_Q + KV_Q(n_t).
\end{equation}
The dominant extra memory terms are the teacher weights and teacher KV cache.
Once TRB is inactive and the teacher state is released, memory usage returns to the student-only generation profile.

From a FLOP perspective, the additional work from trust-region search, binary search over the interpolation coefficient, and log-space blending is small relative to the transformer forward passes.
Ignoring these lower-order vector operations, the teacher-side FLOP count remains the same order as in standard OPD: the teacher is still evaluated once per generated token, but sequentially rather than in a batched pass.

\section{EOS Canonicalization under Tokenizer Mismatch}
\label{app:eos}

The relevant tokenizer mismatch is that EOS is represented by different raw tokens.
Let $e_S$ and $e_T$ denote the student and teacher EOS tokens.
To avoid splitting the same semantic stop event across two coordinates, we map both tokens to a shared event $e_\star$ before sampling or evaluating KL.
For model $M\in\{S,T\}$, let $\phi_M$ map its native EOS token to $e_\star$ and act as the identity elsewhere.
The aligned distribution is
\begin{equation}
\tilde p_M(v \mid h)
=
\sum_{u:\,\phi_M(u)=v} p_M(u \mid h).
\end{equation}
Sampling and sparse-support KL are computed from $\tilde p_S$ and $\tilde p_T$, so the stop event is compared once rather than split between $e_S$ and $e_T$.
In the present setup, each model contributes a single EOS token, so the implementation reduces to moving its EOS probability to $e_\star$, masking the duplicate coordinate, and emitting $e_S$ when the aligned sampler selects $e_\star$.

\section{Derivation of the Trust-Region Solution}
\label{app:effective_objective}

For completeness, we derive the per-prefix trust-region solver introduced in Section~\ref{sec:method}.
Starting from Eq.~\ref{eq:tr-blend}, introduce a Lagrange multiplier $\eta \ge 0$ for the student-centered KL constraint and $\lambda$ for normalization:
\begin{equation}
\begin{split}
\mathcal{L}(\mu,\eta,\lambda)
&=
\sum_a \mu(a)\log\frac{\mu(a)}{\pi_T(a)}
+ \eta \sum_a \mu(a)\log\frac{\mu(a)}{\pi_S(a)}
+ \lambda \left(\sum_a \mu(a)-1\right),
\end{split}
\end{equation}
where additive constants independent of $\mu$ are omitted.
Setting $\partial \mathcal{L}/\partial \mu(a)=0$ yields
\begin{equation*}
(1+\eta)\log \mu(a)
=
\log \pi_T(a)+\eta \log \pi_S(a)+c,
\end{equation*}
for a scalar constant $c$.
Hence
\begin{equation*}
\mu(a)
\propto
\pi_T(a)^{\frac{1}{1+\eta}}
\pi_S(a)^{\frac{\eta}{1+\eta}},
\end{equation*}
or equivalently, with
\[
\beta = \frac{1}{1+\eta},
\qquad
1-\beta = \frac{\eta}{1+\eta},
\]
\begin{equation}
\mu_\beta(a)
\propto
\pi_S(a)^{1-\beta}\pi_T(a)^\beta,
\end{equation}
which is exactly the family in Eq.~\ref{eq:geom-family}.

\paragraph{Why bisection is valid}
The implementation uses binary search to find the largest $\beta \in [0,1]$ such that
\[
D_{\mathrm{KL}}(\mu_\beta \,\|\, \pi_S) \le \varepsilon.
\]
This is valid because the map
\[
\beta \mapsto D_{\mathrm{KL}}(\mu_\beta \,\|\, \pi_S)
\]
is monotone nondecreasing.

Let
\[
p(a)=\pi_S(a \mid h),
\qquad
q(a)=\pi_T(a \mid h),
\]
and define
\[
r(a)=\log q(a)-\log p(a).
\]
Then the trust-region family can be written as
\begin{equation}
\mu_\beta(a)
=
p(a)\exp\!\bigl(\beta r(a)-A(\beta)\bigr),
\end{equation}
where
\begin{equation}
A(\beta)
=
\log \sum_b p(b)\exp\!\bigl(\beta r(b)\bigr)
\end{equation}
is the log-normalizer.

Now compute the KL divergence to the student:
\begin{equation*}
\begin{split}
D_{\mathrm{KL}}(\mu_\beta \,\|\, p)=
\sum_a \mu_\beta(a)\log\frac{\mu_\beta(a)}{p(a)}=
\sum_a \mu_\beta(a)\bigl(\beta r(a)-A(\beta)\bigr)=
\beta\,\mathbb{E}_{a \sim \mu_\beta}[r(a)]-A(\beta).
\end{split}
\end{equation*}
Since
\[
A'(\beta)=\mathbb{E}_{a \sim \mu_\beta}[r(a)],
\]
we obtain
\begin{equation}
D_{\mathrm{KL}}(\mu_\beta \,\|\, p)
=
\beta A'(\beta)-A(\beta).
\end{equation}
Differentiating once more gives
\begin{equation}
\frac{d}{d\beta}
D_{\mathrm{KL}}(\mu_\beta \,\|\, p)
=
\beta A''(\beta).
\end{equation}
Finally,
\begin{equation}
A''(\beta)
=
\mathrm{Var}_{a \sim \mu_\beta}[r(a)]
\ge 0.
\end{equation}
Hence
\[
\frac{d}{d\beta}
D_{\mathrm{KL}}(\mu_\beta \,\|\, p)
=
\beta\,\mathrm{Var}_{a \sim \mu_\beta}[r(a)]
\ge 0,
\]
which proves that $D_{\mathrm{KL}}(\mu_\beta \,\|\, \pi_S)$ is monotone nondecreasing in $\beta$.
Hence the feasible set
\[
\{\beta \in [0,1] : D_{\mathrm{KL}}(\mu_\beta \,\|\, \pi_S) \le \varepsilon\}
\]
is an interval, and the optimal coefficient $\beta^*$ can be found by binary search on $[0,1]$.

\section{Small-Budget Efficiency of Trust Regions}
\label{app:small_budget}

For small trust-region budgets, the blend path has a favorable local trade-off:
moving slightly away from the student yields a first-order reduction in teacher KL while paying only a second-order behavior-KL cost.

Fix a prefix $h$ and write
\[
p(a)=\pi_S(a\mid h),
\qquad
q(a)=\pi_T(a\mid h).
\]
Assume $p$ and $q$ have common support, and define
\[
r(a)=\log q(a)-\log p(a),
\]
\[
\sigma_p^2
=
\mathrm{Var}_{a\sim p}[r(a)].
\]
Assume $\sigma_p^2>0$; the case $\sigma_p^2=0$ is degenerate, since $p$ and $q$ agree on the support up to normalization.

The trust-region path can be written as the exponential tilt
\begin{equation}
\mu_\beta(a)
=
p(a)\exp\bigl(\beta r(a)-A(\beta)\bigr),
\end{equation}
where
\begin{equation}
A(\beta)
=
\log\sum_b p(b)\exp(\beta r(b)).
\end{equation}
Thus
\[
A'(0)=\mathbb{E}_p[r],
\qquad
A''(0)=\sigma_p^2.
\]

The student-centered behavior KL is
\begin{equation}
D_{\mathrm{KL}}(\mu_\beta\|p)
=
\beta A'(\beta)-A(\beta).
\end{equation}
A Taylor expansion around $\beta=0$ gives
\begin{equation}
D_{\mathrm{KL}}(\mu_\beta\|p)
=
\frac{1}{2}\beta^2\sigma_p^2
+
O(\beta^3).
\label{eq:small_beta_student_kl_app}
\end{equation}
Thus the behavior-KL cost of moving away from the student is second-order in $\beta$.

The teacher KL is
\begin{equation}
D_{\mathrm{KL}}(\mu_\beta\|q)
=
(\beta-1)A'(\beta)-A(\beta),
\end{equation}
while
\begin{equation}
D_{\mathrm{KL}}(p\|q)
=
-A'(0).
\end{equation}
Therefore
\begin{equation}
\begin{aligned}
&D_{\mathrm{KL}}(p\|q)
-
D_{\mathrm{KL}}(\mu_\beta\|q)
 =
\beta\sigma_p^2
+
O(\beta^2).
\end{aligned}
\label{eq:small_beta_teacher_gain_app}
\end{equation}
Thus the reduction in KL to the teacher is first-order in $\beta$.

Now suppose the trust-region coefficient is chosen by the active constraint
\[
D_{\mathrm{KL}}(\mu_\beta\|p)=\varepsilon.
\]
From Eq.~\ref{eq:small_beta_student_kl_app},
\begin{equation}
\beta^*(\varepsilon)
=
\sqrt{\frac{2\varepsilon}{\sigma_p^2}}
+
O(\varepsilon).
\label{eq:small_eps_beta_star_app}
\end{equation}
Substituting into Eq.~\ref{eq:small_beta_teacher_gain_app} gives
\begin{equation}
\begin{aligned}
&D_{\mathrm{KL}}(p\|q)
-
D_{\mathrm{KL}}(\mu_{\beta^*}\|q)
 =
\sqrt{2\varepsilon\sigma_p^2}
+
O(\varepsilon).
\end{aligned}
\label{eq:small_eps_teacher_gain_app}
\end{equation}

Thus, at a fixed prefix, a small KL budget buys a teacher-closeness improvement of order $\sqrt{\varepsilon}$ while paying behavior-KL cost $\varepsilon$ to the student.
This is the local sense in which trust-region warmup is efficient: the earliest movement toward the teacher has high marginal value under a student-centered budget.

\section{Sequence-Level Control from Token-Level Trust Regions}
\label{app:sequence_kl}

For notational simplicity, consider a fixed rollout length $T$; the same argument applies to stopped sequences after padding with an absorbing EOS state.
Let the student rollout distribution be
\begin{equation*}
P_S(a_{1:T})
:=
P_S(a_{1:T}\mid x)
=
\prod_{t=1}^T p_t(a_t\mid h_t),
\end{equation*}
and the TRB rollout distribution be
\begin{equation*}
P_{\mu}(a_{1:T})
:=
P_{\mu}(a_{1:T}\mid x)
=
\prod_{t=1}^T \mu_t(a_t\mid h_t),
\end{equation*}
where $h_t=(x,a_{<t})$ is the prefix before step $t$.

\paragraph{Theorem}
Define
\[
\Delta_t(h_t)
\;=\;
D_{\mathrm{KL}}(\mu_t(\cdot\mid h_t)\,\|\,p_t(\cdot\mid h_t)).
\]
If the token-level behavior policy at each prefix is $\mu_t(\cdot\mid h_t)$ and the student policy is $p_t(\cdot\mid h_t)$, then
\begin{equation}
\label{eq:rollout-kl-decomposition}
D_{\mathrm{KL}}(P_{\mu}\,\|\,P_S)
\;=\;
\sum_{t=1}^T
\mathbb{E}_{h_t\sim P_{\mu}}
\bigl[\Delta_t(h_t)\bigr].
\end{equation}
Equivalently,
\[
D_{\mathrm{KL}}(P_{\mu}\,\|\,P_S)
=
\mathbb{E}_{a_{1:T}\sim P_{\mu}}
\left[
\sum_{t=1}^T \Delta_t(h_t)
\right].
\]

As an immediate corollary, if
\[
\Delta_t(h_t)\le \bar\varepsilon_t
\qquad\text{for all } h_t,
\]
then
\begin{equation}
D_{\mathrm{KL}}(P_{\mu}\,\|\,P_S)
\le
\sum_{t=1}^T \bar\varepsilon_t.
\end{equation}
In particular, if the same budget $\bar\varepsilon_t=\varepsilon$ is used at every step, then
\begin{equation}
D_{\mathrm{KL}}(P_{\mu}\,\|\,P_S)
\le
T\varepsilon.
\end{equation}

\paragraph{Proof}
Start from the definition of rollout-level KL:
\begin{equation*}
D_{\mathrm{KL}}(P_{\mu}\,\|\,P_S)
=
\mathbb{E}_{a_{1:T}\sim P_{\mu}}
\left[
\log \frac{P_{\mu}(a_{1:T}\mid x)}{P_S(a_{1:T}\mid x)}
\right].
\end{equation*}
Using the autoregressive factorizations,
\begin{equation*}
\log \frac{P_{\mu}(a_{1:T}\mid x)}{P_S(a_{1:T}\mid x)}
=
\sum_{t=1}^T
\log \frac{\mu_t(a_t\mid h_t)}{p_t(a_t\mid h_t)}.
\end{equation*}
Therefore
\begin{align*}
D_{\mathrm{KL}}(P_{\mu}\,\|\,P_S)
&=
\sum_{t=1}^T
\mathbb{E}_{a_{1:T}\sim P_{\mu}}
\left[
\log \frac{\mu_t(a_t\mid h_t)}{p_t(a_t\mid h_t)}
\right].
\end{align*}
Now define the local log-ratio
\[
\ell_t(a_t,h_t)
=
\log \frac{\mu_t(a_t\mid h_t)}{p_t(a_t\mid h_t)}.
\]
Applying the tower property and conditioning first on the prefix $h_t$ gives
\begin{align*}
&\mathbb{E}_{a_{1:T}\sim P_{\mu}}
\bigl[\ell_t(a_t,h_t)\bigr]
=\mathbb{E}_{h_t\sim P_{\mu}}
\left[
\mathbb{E}_{a_t\sim \mu_t(\cdot\mid h_t)}
\bigl[\ell_t(a_t,h_t)\bigr]
\right] =
\mathbb{E}_{h_t\sim P_{\mu}}
\bigl[\Delta_t(h_t)\bigr].
\end{align*}
Summing over $t$ proves Eq.~\ref{eq:rollout-kl-decomposition}.
The sequence-level upper bound follows immediately by replacing each local KL term with its uniform bound $\bar\varepsilon_t$.

\end{document}